\documentclass[letterpaper, 10 pt, conference]{ieeeconf}  

\IEEEoverridecommandlockouts                              

\usepackage[utf8]{inputenc}
\usepackage[T1]{fontenc}
\usepackage{graphicx}
\usepackage{amsmath}
\usepackage{amssymb}
\usepackage{multicol}
\usepackage{dblfloatfix}

\title{\LARGE \bf
A Fast Integrated Planning and Control Framework for Autonomous Driving via Imitation Learning
}

\author{Liting Sun, Cheng Peng, Wei Zhan and Masayoshi Tomizuka
\thanks{The authors are with the department of Mechanical Engineering,
        University of California, Berkeley, CA, USA, 94720. }
\thanks{Emails: {\{\tt{litingsun}, \tt{chengpeng2014}, wzhan, tomizuka\} @berkeley.edu.}}
    }

\begin{document}
\maketitle
\thispagestyle{empty}
\pagestyle{empty}

\begin{abstract}
For safe and efficient planning and control in autonomous driving, we need a driving policy which can achieve desirable driving quality in long-term horizon with guaranteed safety and feasibility. Optimization-based approaches, such as Model Predictive Control (MPC), can provide such optimal policies, but their computational complexity is generally unacceptable for real-time implementation. To address this problem, we propose a fast integrated planning and control framework that combines learning- and optimization-based approaches in a two-layer hierarchical structure. The first layer, defined as the \textit{policy layer}, is established by a neural network which learns the long-term optimal driving policy generated by MPC. The second layer, called the \textit{execution layer},  is a short-term optimization-based controller that tracks the reference trajecotries given by the \textit{policy layer} with guaranteed short-term safety and feasibility. Moreover, with efficient and highly-representative features, a small-size neural network is sufficient in the \textit{policy layer} to handle many complicated driving scenarios. This renders online imitation learning with Dataset Aggregation (DAgger) so that the performance of the \textit{policy layer} can be improved rapidly and continuously online. Several exampled driving scenarios are demonstrated to verify the effectiveness and efficiency of the proposed framework.
\end{abstract}

\section{INTRODUCTION}
\label{sec:intro}
In recent years, autonomous driving has attracted a great amount of research efforts in both academia and industry for its potential benefits on safety, accessibility, and efficiency. Typically, autonomous driving systems are partitioned into hierarhical structures including perception, decision-making, motion planning and vehicle control (see \cite{buehler20072005, buehler2009darpa}). Among them, planning and control are two core yet challenging problems that are responsible for safety and efficiency. They should 

\begin{enumerate}
	\item comprehensively consider all possible constraints regarding safety and feasibility (kinematically and dynamically) based on the perceived environment information and reasonable prediction of other road participants' behaviors;
	\item generate optimal/near-optimal maneuvers that provide good driving qualities such as smoothness, passengers' comfort and time-efficiency;
	\item solve the problem within limited runtime to timely respond to rapid changes in surrounding environment.
\end{enumerate}

Simultaneously satisfying the above requirements can be difficult and many planning and control approaches have been proposed \cite{gonzalez2016review}. They can be categorized into four groups: i) graph-search-based, such as $A^*$(see \cite{ferguson2008motion}),  ii) sampling-based (e.g., RRT, RRT*, see \cite{kuwata2009real, karaman2010optimal}),  iii) interpolating-curve-based \cite{berglund2010planning, liang2012automatic}, and iv) optimization-based \cite{madaas2013path, li2014unified}.  Groups i) and ii) are discrete in the sense that they discretize the state space into grids/lattices or sampled nodes and search for solutions that build feasible connections among them. Consequently, the resulting paths/trajectories are not continuous but jerky, and a sub-level smoother is necessary for implementation. Approaches in iii) can mostly generate smooth paths, but it is hard to guarantee its optimality, not even locally. Besides, without temporal consideration, dealing with moving obstacles can be time-consuming using these approaches. Optimization-based approaches, on the other hand, provides a unified formulation in continuous state space. They can easily incorporate all possible constraints and customize cost functions to achieve good driving policies. A typical example is Model Predictive Control (MPC). With the scheme of re-planning, a constrained optimization problem is solved at each time step so that the optimal control actions (accelerations and steering angles) are directly generated to drive the ego-vehicles. However, to guarantee persistent feasibility and safety, a long horizon is commonly preferred, which might fail MPC for real-time applications in terms of requirement 3).

A common strategy to address this problem is to manually terminate the optimization process once the pre-determined runtime is exhausted. For instance, in \cite{ziegler2014trajectory}, with a horizon length of $N{=}30$, iteration of the optimizer was terminated after $0.5$s. Obviously, such strategy forces realtimeness at the cost of optimality, and consequently a warm start is necessary to achieve good performance. Hence, a natural question arises:

\textit{Can we create a policy that generates outputs within the pre-determined runtime, and mimics the optimal driving policy provided by long-term MPC in terms of driving qualities, feasibility and safety?}

Motivated by this, we propose a two-layer hierarchical structure that combines learning- and optimization-based methods. The first layer, defined as the \textbf{\textit{policy layer}}, is constructed by a neural network that takes in current perception information and generates instructive trajectories by learning the long-term optimal and safe driving policy given by MPC. The second layer, called the \textbf{\textit{execution layer}}, tracks the instructive trajectories with further guaranteed short-term feasibility and safety by solving a short-horizon optimization problem. Via learning, such an architecture saves the effort to solve the most time-consuming optimization problem in long-term MPC to satisfy the runtime requirement. Moreover, to continuously improve the fidelity of the \textbf{\textit{policy layer}}, a customized DAgger approach, defined as Sampled-DAgger, is developed. The \textbf{\textit{policy layer}}, initially trained on a training set $\mathcal{D}_0$, is tested continuously for different driving scenarios with the expert MPC running in parallel at a slow rate, so that sampled failure cases will be labelled\footnote[1]{Label: the process to query for the optimal solution for a given state configuration. Specifically in this work, it means solving a long-horizon MPC problem that considers long-term feasibility, safety and efficiency.} with the expert policy to generate a new incremental training set $\mathcal{D}'$. By iteratively aggregating the training set ($\mathcal{D}{=}\mathcal{D}{\cup}\mathcal{D}'$) and re-training the neural network, the \textbf{\textit{policy layer}} will converge to the imitated optimal policy (long-term MPC) with high fidelity.

In autonomous driving, many learning-based control approaches have been investigated. A large community focuses on the End-to-End learning that directly constructs a mapping from the sensory inputs (such as images) to driving actions via neural networks (for instance, see \cite{pomerleau1989alvinn,bojarski2016end}). Such End-to-End approaches, however, suffer from several drawbacks. First, as pointed by \cite{chen2015deepdriving}, the decision-making level of the neural networks is too low to assure feasibility, safety and smoothness. Second, the End-to-End approaches typically learn from human drivers, which makes it hard to guarantee the optimality of the imitated driving policy since different human drivers have different preferences to even similar scenarios. Finally, the sensory inputs are usually images that contain lots of redundant information with respect to the low-level commands. Thus, deep convolutional neural networks (CNN) with multiple layers are preferred to extract useful features and achieve good training performance. Such complicated network structures require long training time (hours/days of training on GPUs), which makes it impractical to be improved with online DAgger. This can be dangerous in practice since any real-time deviation from the training set can lead to unpredictable error propagation and cause severe consequences. 

On the contrary, our proposed planning and control framework utilizes ``middle-level'' learning that benefits from:
\begin{itemize}
	\item ``middle-level'' learning-based decision variables (i.e., trajectories) that leave margins for the \textbf{\textit{execution layer}} to further guarantee short-term feasibility and safety.
	\item ``middle-level'' sensory inputs. Instead of raw images, we extract highly representative features that significantly reduce the complexity of the neural network. A single-hidden-layer static network with minutes of training time on CPU is sufficient. Such structural simplicity and time-efficient training render online DAgger so that the performance of the \textbf{\textit{policy layer}} can be rapidly and continuously improved. As a result, robustness and adaptability of the system are enhanced.
\end{itemize}
Moreover, our proposed framework imitates the long-term driving policy given by MPC instead of human drivers. This further speeds up the learning process by avoiding possible confusion caused by the non-optimality and individual differences of the human drivers' data.

The remainder of this paper is organized as follows. Section \ref{sec:preliminaries} presents the overall structure with the proposed framework, as well as the formulation of the long-term expert MPC for imitation in \textbf{\textit{policy layer}} and the short-term optimization in \textbf{\textit{execution layer}}. Section \ref{sec:rdl} illustrates the construction of the \textbf{\textit{policy layer}} with details on feature extraction and training process through DAgger. Section \ref{sec:generalization} generalizes the abstract training frames to more complicated driving scenarios. Illustrative examples are given in Section \ref{sec:results} and Section \ref{sec:conclusion} concludes the paper. A complementary video can be found at: \textit{http://iros2017autodrivinglearning.weebly.com}.

\section{PRELIMINARIES}
\label{sec:preliminaries}
\subsection{Overview of the Hierarchical Structure}
\begin{figure}[h]
	\begin{centering}
		\includegraphics[clip,scale=0.35]{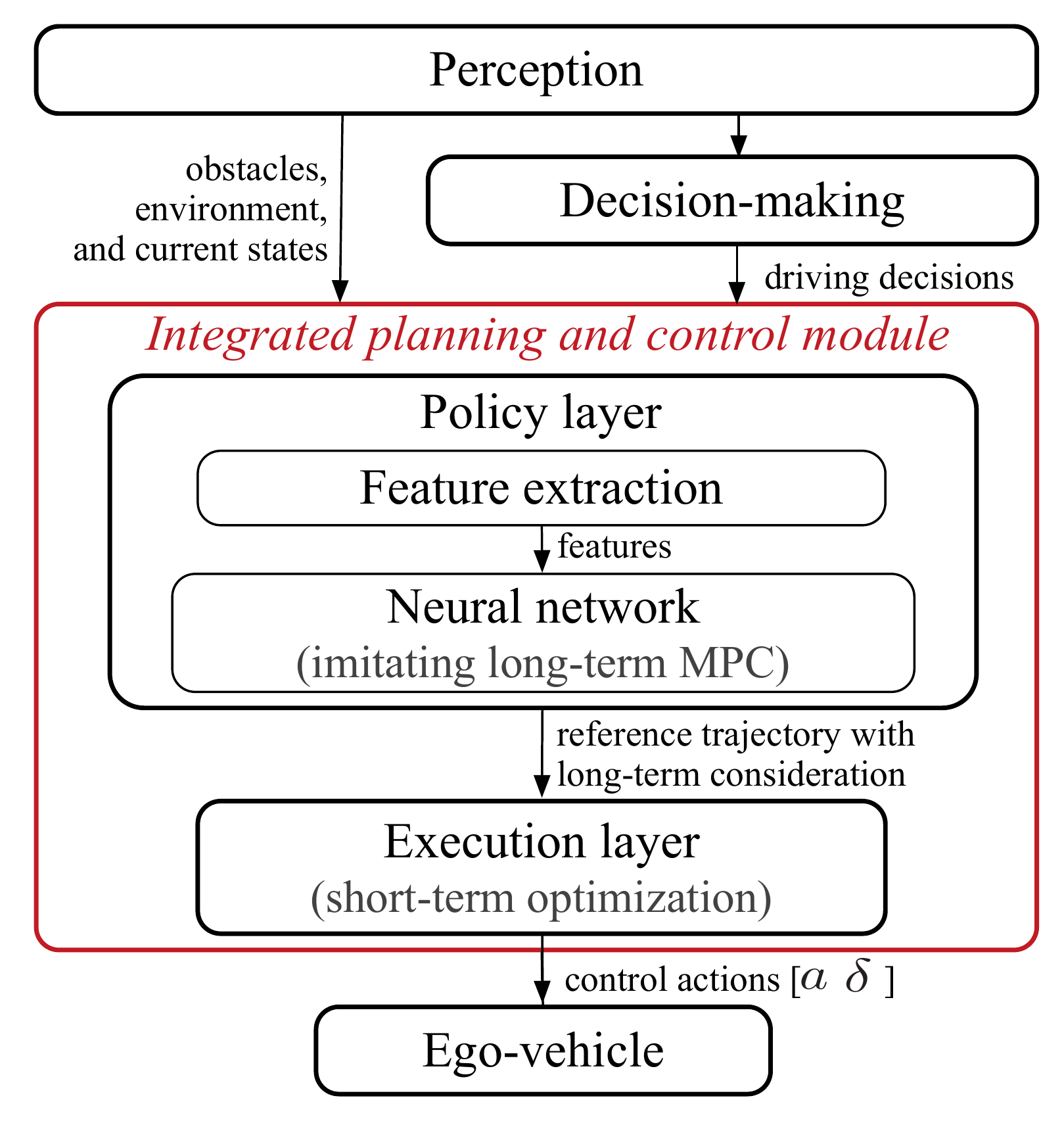}
		\par
	\end{centering}
	\caption{The overall hierarchical structure \label{fig:overall_structure}}
\end{figure}

Figure \ref{fig:overall_structure} shows the overview of the hierarchical structure with the proposed fast integrated planning and control framework for autonomous driving. It consists of three modules: perception, decision-making, and planning and control. At each time step, the perception module detects the surrounding environment, and yields measurements/estimates of all necessary states of the ego-vehicle. For instance, the location, orientation, velocity and its relative positions and velocities to all other visible road participates (static or moving) can be detected or estimated. Based on all perceived information and pre-defined driving tasks, the decision-making module will set the reference lane and horizon goal correspondingly to instruct the next-level planning and control.

In this paper, we focus on the planning and control module. To satisfy all the three requirements discussed in Section \ref{sec:intro}, the planning and control module we proposed also employs a hierarchical structure. The first layer is the \textbf{\textit{policy layer}} that includes feature extraction and a neural network. Based on driving decisions, the feature extraction converts the redundant and complicated perception output into an abstract driving scenario that can be effectively described by a group of highly-representative features. Then the features are fed into the neural network to fastly yield reference trajectories that imitate the optimal trajectories provided by long-term MPC. Finally, such reference trajectories are given to the \textbf{\textit{execuation layer}}, which generates corresponding control actions by solving a short-horizon optimization problem to further guarantee short-term feasibility and safety.

Due to its learning-based characteristic, the proposed planning and control framework can respond extremely fast while achieving similar long-term smoothness and safety with the expert MPC policy. 

\subsection{Long-Term Model Predictive Control}
\begin{figure}[h]
	\begin{centering}
		\includegraphics[clip,scale=0.5]{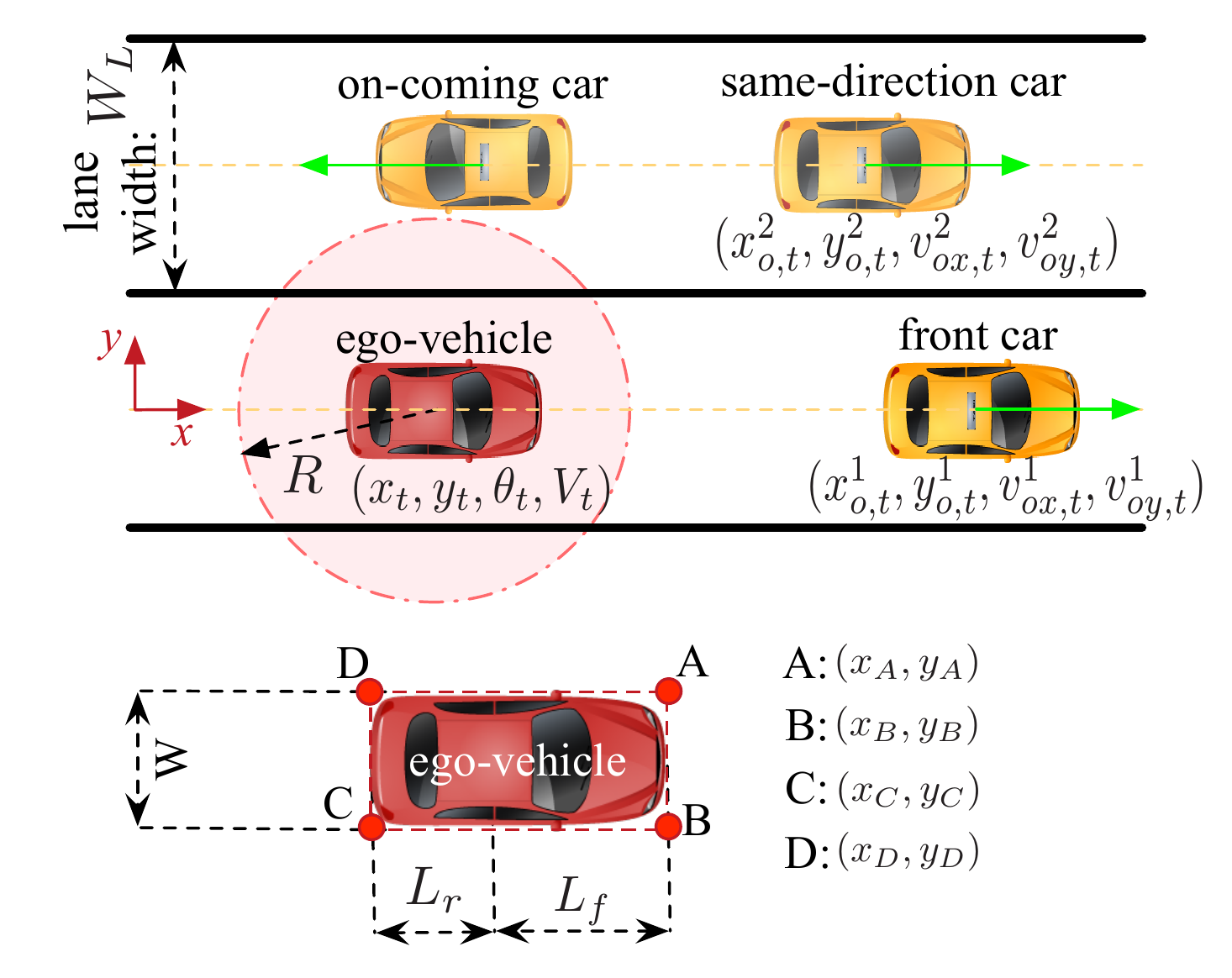}
		\par
	\end{centering}
	\caption{The abstract scenario for long-term MPC formulation \label{fig:long_MPC_scenario}}
\end{figure}

In this section, the formulation of the expert long-term MPC is presented. As shown in Fig.~\ref{fig:long_MPC_scenario}, we consider the abstract driving scenario with two lanes, where there is one front car and another car on the adjacent lane that is either on-coming or with the same direction (depending on the real environment, direction of the adjacent lane is fixed). At each time instant $t$, states of the ego-vehicle and surrounding cars are denoted, respectively, by $z_t{=}[x_t,y_t,$ yaw angle $\theta_t$, speed $V_t]^T$ and $(x_{o,t}^i, y_{o,t}^i,v_{ox,t}^i,v_{oy,t}^i)$ ($i{=}1$ for the front car and $i{=}2$ for the adjacent-lane car). Control actions of the ego-vehicle are the longitudinal acceleration and steering angle, denoted by $u_t{=}[a_t,\delta_t]^T$. 

Define the preview horizon of MPC by $N$ ($N{=}30$ is set in this work). Within the horizon, the predicted ego-vehicle's states and corresponding control variables are respectively represented by $z^p_k{=}[x^p_k,y^p_k,\theta^p_k,V^p_k]^T$ with $k{=}0,1,2,{\cdots}, N$ and $u^p_k{=}[a^p_k,\delta^p_k]^T$ with $k{=}0,1,2,{\cdots}, N{-}1$. Similarly, the within-horizon surrounding cars' states are denoted by $(x_{o,k}^{p,i}, y_{o,k}^{p,k},v_{ox,k}^{p,i},v_{oy,k}^{p,i})$ for $i{=}1,2$ and $k{=}0,1,2,{\cdots}, N$. Moreover, it is assumed that the surrounding cars move constantly at its current speed within the horizon, i.e., $v^{p,i}_{ox,k}{=}v^{i}_{ox,t}$ for $k{=}0,1,2,{\cdots}, N$. 

\subsubsection{Objective Function}
To generate smooth and efficient driving policy, the long-term MPC penalizes the ego-vehicle to drive as fast as the speed limit allows. Hence, the cost function is defined as:
\begin{equation}
	\label{eq:long_MPC_cost_function}
	J_t=\sum_{k=0}^N \left[
	\begin{aligned}
	x^e_k\\
	y^e_k
	\end{aligned}
	\right]^T W_e\left[
	\begin{aligned}
	x^e_k\\
	y^e_k
	\end{aligned}
	\right]+\sum_{k=0}^{N{-}1}(\triangle u^p_k)^TW_u\triangle u^p_k
\end{equation}
where $[x^e_k; y^e_k]{=}[x^p_k{-}x_t^{goal};y^p_k{-}y_t^{goal}]$ represents the distances between the predicted positions $(x^p_k, y^p_k)$ and the given goal position $(x_t^{goal},y_t^{goal})$ that is set ahead of the vehicle. $\triangle u^p_k{=}u^p_k{-}u^p_{k-1}$ (for $k{=}0$, $\triangle u^p_0{=}u^p_0{-}u_{t{-}1}$) is the change of the control variables between two sucessive operations within the horizon. Both $W_e$ and $W_u$ are positive definite, i.e., $W_e{\in}\mathcal{R}^{2\times2}{>}0,W_u{\in}\mathcal{R}^{2\times2}{>}0$. The decision variables include both states and control variables, namely, $(z^p_0,z^p_1,{\cdots},z^p_N, u^p_0, u^p_1, {\cdots}, u^p_{N{-}1})$.
\subsubsection{Constraints}
The long-term MPC comprehensively considers all constraints from the system kinematics feasibility (equality constraints), dynamics feasibility and collision-free safety (inequality constraints). 

Regarding to the kinematics feasibility, Bicycle model \cite{rajamani2011vehicle} is adopted, which means that the decision variables (states and control variables) have to satisfy the defined manifold within the horizon, i.e., for $k{=}0,1,{\cdots},N{-}1$,
\begin{IEEEeqnarray}{rCl}
	x^p_{k{+}1}&=&x^p_k+V^p_k\cos\left(\theta^p_k{+}\tan^{-1}\left(\frac{L_r}{L}\tan\delta^p_k\right)\right)dt	\label{eq:Bicycle_Model_1}\\
	y^p_{k{+}1}&=&y^p_k+V^p_k\sin\left(\theta^p_k{+}\tan^{-1}\left(\frac{L_r}{L}\tan\delta^p_k\right)\right)dt\\
	\theta^p_{k{+}1}&=&\theta^p_k+V^p_k\frac{\tan\delta^p_k}{L}\cos\left(\tan^{-1}\left(\frac{L_r}{L}\tan\delta^p_k\right)\right)dt\quad\\
	V^p_{k{+}1}&=&V^p_k+a^p_kdt	\label{eq:Bicycle_Model_4}
\end{IEEEeqnarray}
and $x^p_0{=}x_t, y^p_0{=}y_t, \theta^p_0{=}\theta_t, V^p_0{=}V_t$.
$L{=}L_r{+}L_f$ is the total car length as shown in Fig.~\ref{fig:long_MPC_scenario} and $dt{=0.1}$s is the discrete time interval.

The dynamics feasibility is guaranteed via constraints from the G-G diagram. \begin{figure}[h]
	\begin{centering}
		\includegraphics[clip,scale=0.35]{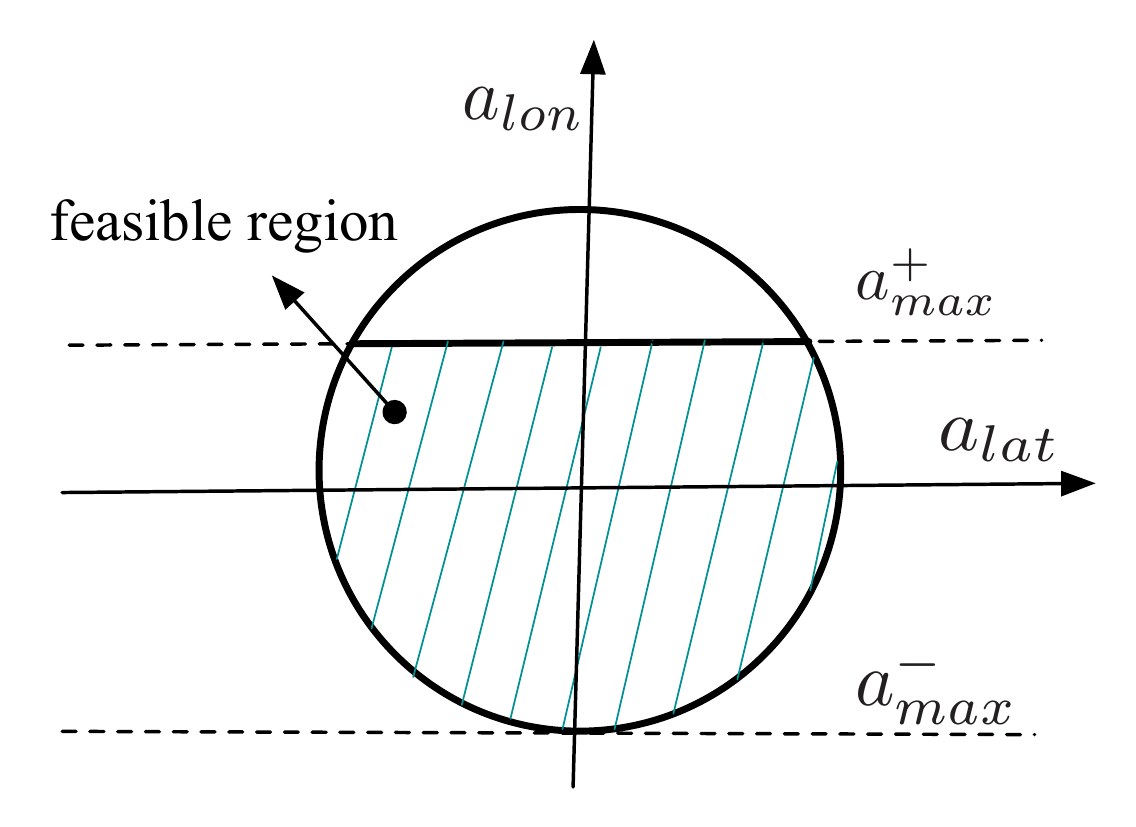}
		\par
	\end{centering}
	\caption{Acceleration boundaries from G-G diagram for dynamics feasibility\label{fig:dynamic_boundary}}
\end{figure}
As shown in Fig.~\ref{fig:dynamic_boundary}, $a^+_{max}$ and $a^-_{max}$ represent, respectively, the maximum acceleration input and deceleration input, and the longitudinal and lateral accelerations $a_{lon}$ and $a_{lat}$ have to sit within the feasible region. 
Let
\begin{IEEEeqnarray}{rCl}
	a^p_{x,k}&=&\dfrac{x^p_k{-}2x^p_{k{-}1}{+}x^p_{k{-}2}}{dt^2}\label{eq:a_x}\\
	a^p_{y,k}&=&\dfrac{y^p_k{-}2y^p_{k{-}1}{+}y^p_{k{-}2}}{dt^2}\label{eq:a_y}
\end{IEEEeqnarray}
be the $x$-/$y$-direction accelerations and notice that $a^2_{lon}{+}a^2_{lat}{=}a^2_x{+}a^2_y$, hence the feasible region in the G-G diagram can be expressed by:
\begin{IEEEeqnarray}{lll}
	&&(a^p_{x,k})^2{+}(a^p_{y,k})^2{\le}(a^{-}_{max})^2\label{eq:dynamic_constraint_1}\\
	&&a^p_k{\triangleq} a^p_{lon,k}\le a^{+}_{max}, \forall k{=}0,1,{\cdots},N{-}1.\label{eq:dynamic_constraint_2}
\end{IEEEeqnarray}

Finally, safety is assured by constrainting the ego-vehicle's configurations (position and orientation) and its distances to surrounding cars. As shown in Fig.~\ref{fig:long_MPC_scenario}, assuming $\epsilon$ is the safety buffer distance between two cars, we define the safe distance as $R=L_r{+}L_f{+}\epsilon$ and 
\begin{equation}
d^i_k=\sqrt{(x^p_k{-}x^{p,i}_{o,k})^2{+}(y^p_k{-}y^{p,i}_{o,k})^2}\ge R,i{=}1,2\label{eq: safe_R}
\end{equation}
is required throughout the horizon. Recalling the fact that we assume all surrounding vehicles move constantly within the horizon, therefore $x^{p,i}_{o,k}{=}x^{i}_{o,t}{+}kv^i_{ox,t}dt$ and $y^{p,i}_{o,k}=y^{i}_{o,t}+kv^i_{oy,t}dt$. 

Moreover, to avoid collisions with static obstacles such as curb, the following constraints on the ego-vehicle's configurations are included (refer to Fig.~\ref{fig:long_MPC_scenario} for definition of points \textit{A}, \textit{B}, \textit{C} and \textit{D}):
\begin{IEEEeqnarray}{rCl}
	&&0\le y^p_k\le W_L\label{eq: y_constraints}\\
	&&y_A=x^p_k{+}L_f{\sin}\theta^p_k{+}\frac{W}{2}{\cos}\theta^p_k	\le y_{max}{\triangleq}1.5W_L\label{eq: yA_constraints}\\
	&&y_D=x^p_k{-}L_f{\sin}\theta^p_k{+}\frac{W}{2}{\cos}\theta^p_k	\le y_{max}{\triangleq}1.5W_L\label{eq: yD_constraints}\\
	&&y_B=x^p_k{+}L_f{\sin}\theta^p_k{-}\frac{W}{2}{\cos}\theta^p_k	\ge y_{min}{\triangleq}{-}0.5W_L\label{eq: yB_constraints}\\
	&&y_C=x^p_k{-}L_f{\sin}\theta^p_k{-}\frac{W}{2}{\cos}\theta^p_k	\ge y_{min}{\triangleq}{-}0.5W_L\label{eq: yC_constraints}
\end{IEEEeqnarray}

In the end, physical constraints on the steering angle is imposed, i.e.,
\begin{equation}
\delta_{min}\le\delta^p_k\le\delta_{max}.\label{eq:steering_angle_constraints}
\end{equation}

In summary, the long-term MPC aims to solve the following highly-nonlinear optimization problem at each time instant $t$:
\begin{IEEEeqnarray}{CCC}
	&P:=\min_{(z^p_0,z^p_1,{\cdots},z^p_N,u^p_0,u^p_1,{\cdots},u^p_{N{-}1})} &J_t\label{eq:minization_J}\\
	&s.t& (\ref{eq:Bicycle_Model_1})\text{ - }(\ref{eq:steering_angle_constraints})\nonumber
\end{IEEEeqnarray}
and generate a long-horizon optimal full-state trajectory $(z^{p*}_0,z^{p*}_1,{\cdots},z^{p*}_N)$ as well as the corresponding control actions $(u^{p*}_0,u^{p*}_1,{\cdots},u^{p*}_{N{-}1})$. Note that the optimal solutions of (\ref{eq:minization_J}) aims to drive the ego-vehicle as fast as possible to the goal positions set by the high-level decision-making module, as long as it is safe and feasible. For instance, if the front car is slow and the adjacent lane is clear, the ego-vehicle will automatically overtake the front car. Contrarily, if the front car is fast enough (running at speed limit) or the adjacent lane is occupied by other cars, the ego-vehicle will follow the front car.

\subsection{Short-Term Optimization in the Execution Layer}
The short-term optimization problem in the \textbf{\textit{execution layer}} is formulated similarly to the long-term MPC discussed above, except for two aspects:
\begin{enumerate}
	\item The horizon length is much shorter. $N_{short}{=}5$ is selected to reduce the computation load. 
	\item Instead of penalizing the ego-vehicle to drive to the goal point $(x_t^{goal},y_t^{goal})$ as fast as possible, the \textbf{\textit{execution layer}} tracks the reference trajectory given by the \textbf{\textit{policy layer}}. Suppose the reference trajectory is $[(x^r_1,y^r_1), (x^r_2,y^r_2), (x^r_3,y^r_3), (x^r_4,y^r_4), (x^r_5,y^r_5)]$, then the cost function becomes:
	\begin{equation}
	J^{short}_{t}{=}\sum_{k{=}0}^{N_{short}}\left[
	\begin{aligned}
	x^e_k\\
	y^e_k
	\end{aligned}
	\right]^T W_e\left[
	\begin{aligned}
	x^e_k\\
	y^e_k
	\end{aligned}
	\right]+(\triangle u^{p}_k)^TW_u \triangle u^{p}_k\label{eq:short_cost_function}
	\end{equation}
	where $[x^e_k, y^e_k]{\triangleq}[x^p_k{-}x^r_k,y^p_k{-}y^r_k]$ represents the tracking error.
\end{enumerate}
\section{IMITATION LEARNING OF THE LONG-TERM OPTIMAL DRIVING POLICY}
\label{sec:rdl}
The optimization problem with long-term MPC in (\ref{eq:minization_J}) is generally too time-consuming for real-time applications due to its nonlinearities with both equality and inequality constraints. To address the time efficiency problem while maintaining its optimality (to some extent), a learning-based \textbf{\textit{policy layer}} is constructed to imitate the optimal policy given by (\ref{eq:minization_J}), as discussed in Section \ref{sec:preliminaries}-A. In this section, details with respect to the learning process will be covered.

\subsection{Feature Extraction}
As mentioned above, the perceived information from perception module is redundant and complicated. To render good learning performance, feature selection is of key importantance. Recalling the abstract scenario shown in Fig.~\ref{fig:long_MPC_scenario}, we need a group of features that can represent the following variables at each time instant $t$:
\begin{itemize}
	\item \textit{on-map motion variables}: ego-vehicle's current location and speed within the abstract map settings;
	\item \textit{goal variables}: ego-vehicle's distance to the goal point at current time instant;
	\item \textit{safety variables}: ego-vehicle's current relative positions and velocities with respect to surrounding obstacles/cars.
\end{itemize}

\begin{figure}[h!]
	\begin{centering}
		\includegraphics[clip,scale=0.5]{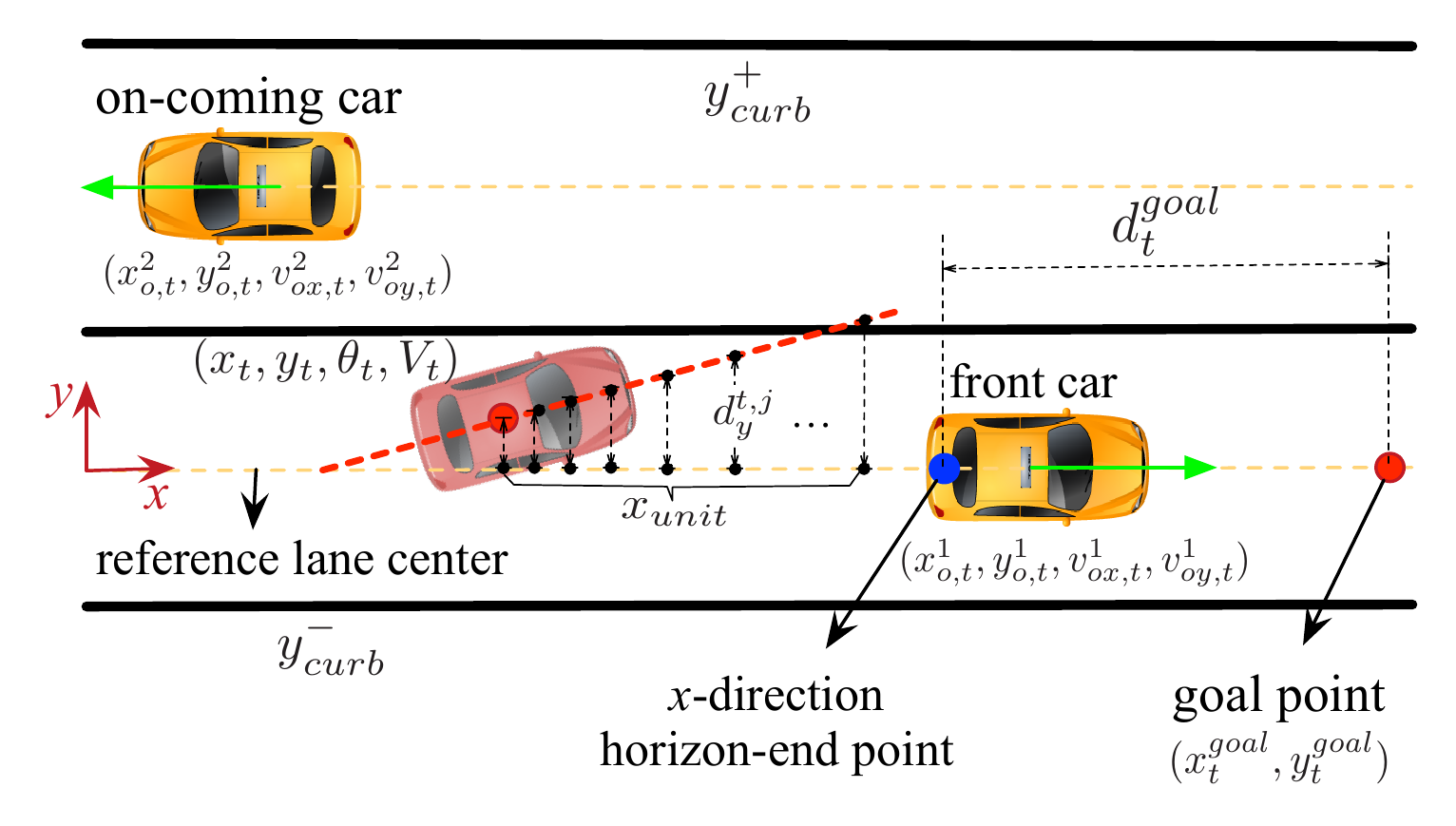}
		\par
	\end{centering}
	\caption{Definitions for feature selection\label{fig:spatial_features}}
\end{figure}

To effectively describe the \textit{on-map motion variables}, first the distances of the ego-vehicle to curbs are given, namely,  $f_t^{map,1}{\triangleq}y_t{-}y^+_{curb}$ and $f_t^{map,2}{\triangleq}y_t{-}y^-_{curb}$. Furthermore, we spatially discretize the proximate reference lane center, and define it as $x_{unit}\triangleq[\triangle x^c_1,\triangle x^c_2,{\cdots},\triangle x^c_j,{\cdots},\triangle x^c_m]$, as shown in Fig.~\ref{fig:spatial_features}. Then the current deviations of the ego-vehicle from the reference lane center can be represented by $d_t^y\triangleq[d_{t}^{y,1},d_{t}^{y,,2},{\cdots},d_{t}^{y,j},{\cdots},d_{t}^{y,m}]$, where each element is given by
\begin{equation}
	\label{eq:definition_dy}
	d_{t}^{y,j}=y_t+\triangle x^c_j \tan\theta_t, \forall j=1,2,\cdots,m.
\end{equation}
In this work, $m{=}10$ is set, generating ten more on-map-motion-related features $f_t^{map,j{+}2} \triangleq d^{t,j}_y$ for $j{=}1,2,{\cdots}, 10$. As for the speed-related feature, it should indicate the current ego-vehicle's speed $V_t$ as well as its margin to the pre-defined speed limit. Hence, we define $f_t^{map,13}\triangleq V_t^{mar}{=}V_t{-}V_{max}$. 

{\textit{Remark I}}: $x_{unit}$ is not necessarily evenly sampled. Actually, noting the fact that closer proximity matters more, $x_{unit}$ can be sampled spatially with increasing intervals, i.e., $0{=}\triangle x^c_1<\triangle x^c_2{-}\triangle x^c_1<{\cdots}<\triangle x^c_m{-}\triangle x^c_{m{-}1}$. Such configuration allows $x_{unit}$ to cover a longer distance, which helps improve the features' sensitivity to small yaw angle changes.

Features on the \textit{goal variables} are extracted in the way that they can drive the ego-vehicle to the goal point as fast as possible.
To realize this, the goal-related features should significantly differ for slow, far-to-goal situations and fast, close-to-goal situations. Therefore, we define the goal-related feature as the $x$-direction distance between the predicted horizon-end position and the goal point, as depicted in Fig.~\ref{fig:spatial_features}. Mathmatically, it is given by
\begin{equation}
\label{eq: }
	f_t^{goal}\triangleq d_t^{goal}=x_t+v_{t,x}Ndt-x_t^{goal},
\end{equation}
where $v_{t,x}{=}V_t\cos\left(\theta_t{+}\tan^{-1}\left(\frac{L_r}{L}\tan\delta_{pre}\right)\right)$ is the $x$-direction vehicle velocity with $\delta_{pre}$ as the previous steering angle input.

Finally, to address the \textit{safety variables}, features on the surrounding vehicles's states from the ego-vehicle's view are included. Since we assume that throughout the long-term horizon, both surrounding vehicles (i{=}1,2) move constantly with their corresponding velocities $(v^{i,t}_{ox},v^{i,t}_{oy})$ at time instant $t$, the following features are extracted, for $i{=}1,2$,
\begin{equation}
	\label{eq: feature_safety}
	f_{t,i}^{safety} {=} \left[x_t{-}x^{i}_{o,t},y_t{-}y^{i}_{o,t},v_{t,x}{-}v^{i}_{ox,t},v_{t,y}{-}v^{i}_{oy,t}\right]^T
\end{equation}

All above defined features $f_t^{map}$, $f_t^{goal}$ and $f_t^{safety}$ generate a set of highly representative features $f_t{=}[(f_t^{map})^T,f_t^{goal}, (f_t^{safety})^T]^T$ and will be fed into the neural network to fastly generate trajecotries that considers long-term smoothness and safety.
\subsection{Structure of the Policy Layer}
The \textbf{\textit{policy layer}} is constructed by a single-hidden-layer neural network whose input nodes are the extracted features $f^t$ selected in Section \ref{sec:rdl}-A. 

\begin{figure}[h]
	\begin{centering}
		\includegraphics[clip,scale=0.5]{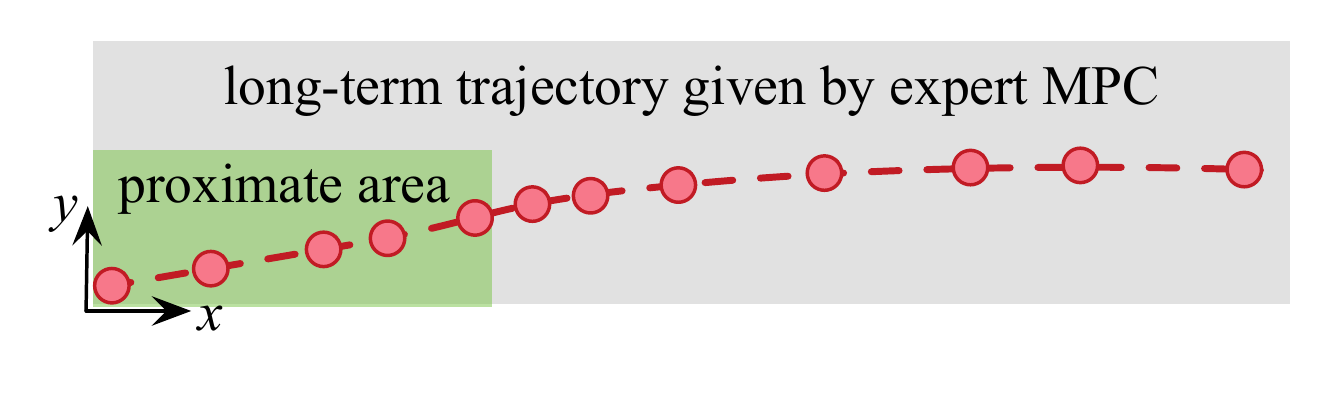}
		\par
	\end{centering}
	\caption{Illustration of the proximate learning concept\label{fig:proximate_learning}}
\end{figure}

As for the output layer, motivated by the first-action-only principle of receding horizon control, we formulate the neural network to put more effort on learning the most proximate trajectories given by the optimal long-term MPC defined by (\ref{eq:minization_J}), as illustrated in Fig.~\ref{fig:proximate_learning}. Such configuration brings two benefits. First, instead of full-horizon trajectory learning, proximate trajectory learning helps reduce the number of nodes in the output layer of the neural network and consequently the network size, which significantly reduces the training time cost. Second, compared to full-horizon learning, proximate trajectory learning generates only short-term reference trajectories for the succeeding \textbf{\textit{execution layer}} instead of long ones. This enables the formulation and fast solution of a short-horizon optimization problem in the \textbf{\textit{execution layer}}, as defined in (\ref{eq:short_cost_function}).

In this work, as mentioned in Section \ref{sec:preliminaries}-C, $N_{short}{=}5$ is set in the \textbf{\textit{execution layer}}. Hence, the output layer of the network consists of ten nodes, namely, $[(x_1, y_1),(x_2, y_2),(x_3, y_3),(x_4, y_4),(x_5, y_5)]$.
\subsection{Learning via Sampled-DAgger Process}
It is well known that, to achieve a good learned policy, supervised learning by itself is often not sufficient since it is never practical to generate a training set that satisfies the same states distribution as the real-world data. Such distribution mismatch may make the learned policy biased and cause severe problems during execution \cite{ross2011reduction}. Once the encountered states slightly deviate from the training set (i.e., unfamilar features), the learned policy might yield wrong output, and such error will propogate and eventually fails the system.

One effective way to solve this is to use imitation learning with DAgger (Dataset Aggregation) \cite{ross2011reduction}. As an iterative algorithm, it adds all on-policy data into the training set so that, over iterations, the training set can cover most scenarios that the learned policy might encounter based on previous experience. 

\begin{figure}[h]
	\begin{centering}
		\includegraphics[clip,scale=0.45]{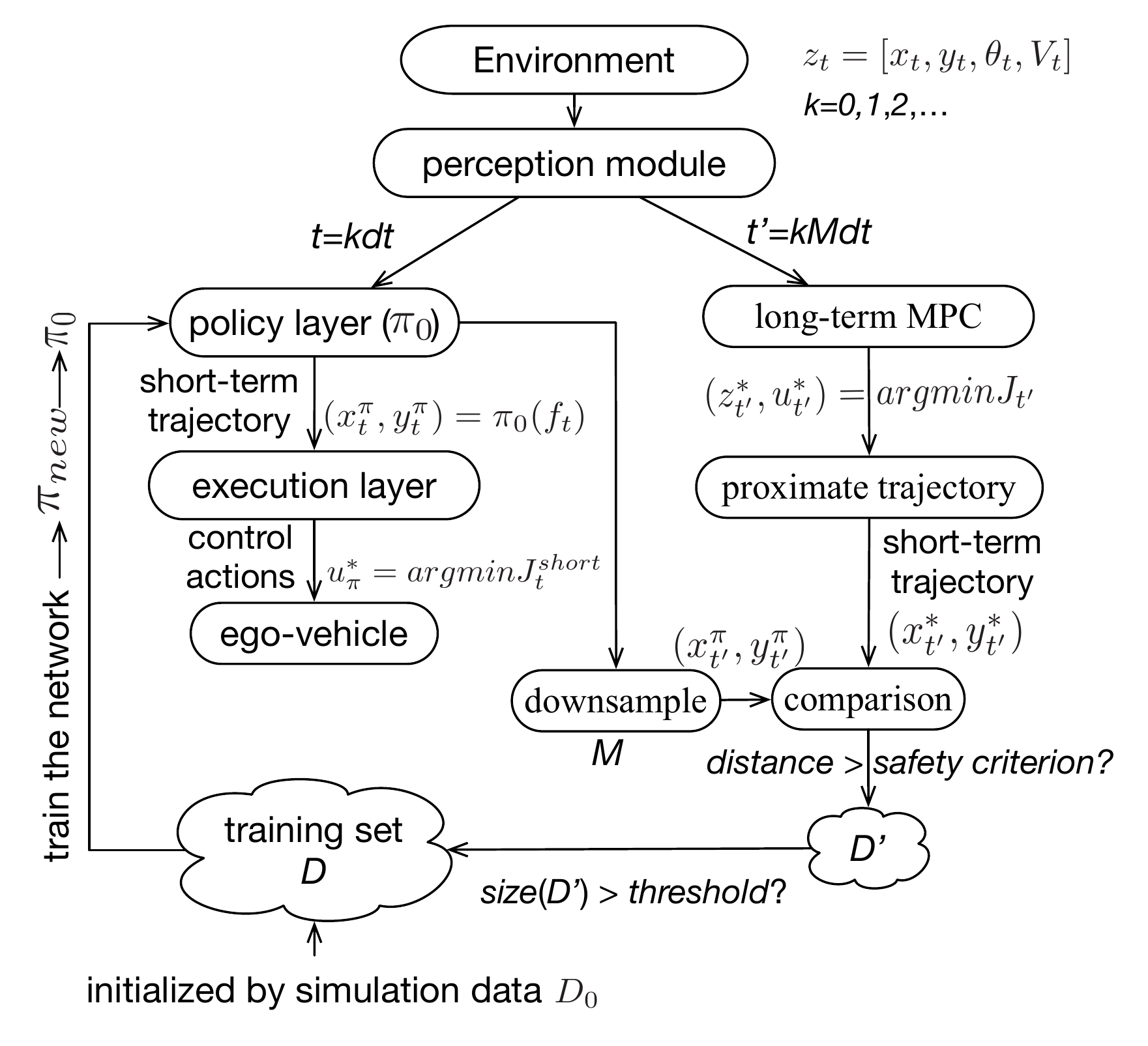}
		\par
	\end{centering}
	\caption{The Sampled-DAgger procedure\label{fig:DAgger_procedure}}
\end{figure}

Motivated by this, a customized DAgger, defined as Sampled-DAgger, is proposed to improve the performance of the \textbf{\textit{policy layer}}. As shown in Fig.~\ref{fig:DAgger_procedure}, the details of the Sampled-DAgger are given as follows:
\begin{enumerate}
	\item Gather an initial training set $\mathcal{D}_0$ by running long-term expert MPC for randomly generated scenarios in simulation and train the network to yield an initial policy $\pi_0$. Initially, around $20k$ scenarios are included in $\mathcal{D}_0$.
	\item Run the autonomous driving architecture shown in Fig.~\ref{fig:overall_structure} with the learned policy $\pi_0$ in \textbf{\textit{policy layer}}. Meanwhile, run the long-term expert MPC in parallel at a slow rate to periodically label the features with optimal outputs. As shown in Fig.~\ref{fig:DAgger_procedure}, the long-term MPC is running every $M$ time steps, i.e., $T^{MPC}_{interval}{=}Mdt$, which is set to be long enough to solve the optimization problem in (\ref{eq:minization_J}).
	\item Compare the optimal proximate trajectory given by $\pi^*(f_{t{=}kT^{MPC}_{interval}})$ with the on-policy trajectory given by $\pi_0(f_{t{=}kT^{MPC}_{interval}})$ for $k{=}0,1, 2, {\cdots}$. If the normalized Euclidean distance between the two trajectories is larger than the pre-defined safety criterion, label the features $f_{t{=}kT^{MPC}_{interval}}$ with corresponding optimal outputs and push them into a new training set $\mathcal{D}'$. Once the size of $\mathcal{D}'$ reaches a pre-defined threshold, go to step 4).
	\item Aggregate previous training set with $\mathcal{D}'$, i.e., $\mathcal{D}=\mathcal{D}\cup\mathcal{D}'$ and re-train the network to yield a new policy $\pi_{new}$ and set $\pi_{0}{=}\pi_{new}$.
	\item Repeat 2) to 4) until the learned policy $\pi_0$ converges.
\end{enumerate}

Compared to the orignial DAgger, the Sampled-DAgger is data-efficient since instead of querying the expert MPC for all on-policy data, it only queries for those with critically deviated outputs. A similar idea to this is SafeDAgger \cite{zhang2016query} which focuses on automatically learning an in-loop safety policy to check which on-policy data should be queried for. While the proposed Sampled-DAgger utilizes pre-defined safety criterion, but adopts the additional downsampling step to allow online queries for the time-consuming expert MPC policy, which helps reduce the offline relabeling effort.

Benefiting from the group of highly-representative features selected in Section \ref{sec:rdl}-A and the data-efficient online aggregation, the Sampled-DAgger is iteratively time-efficient. It only takes 2-3 minutes to train such a small-size network with CPU only, which means that the learned policy in the \textbf{\textit{policy layer}} can be updated rapidly and continuously (with an interval of 2-3 minutes) to make the system more adaptive to environment.
\section{GENERALIZATION TO MORE COMPLICATED SCENARIOS}
\label{sec:generalization}
In Section \ref{sec:preliminaries} and \ref{sec:rdl}, an abstract scenario (Fig.~\ref{fig:long_MPC_scenario}) is formulated to train the \textbf{\textit{policy layer}}. The application of the learned policy, however, is not limited to such simple scenarios. When combined with the decision-making module, the learned policy can be generalized to handle much more practical and complicated driving scenarios.

\begin{figure}[h!]
	\begin{centering}
		\includegraphics[clip,scale=0.32]{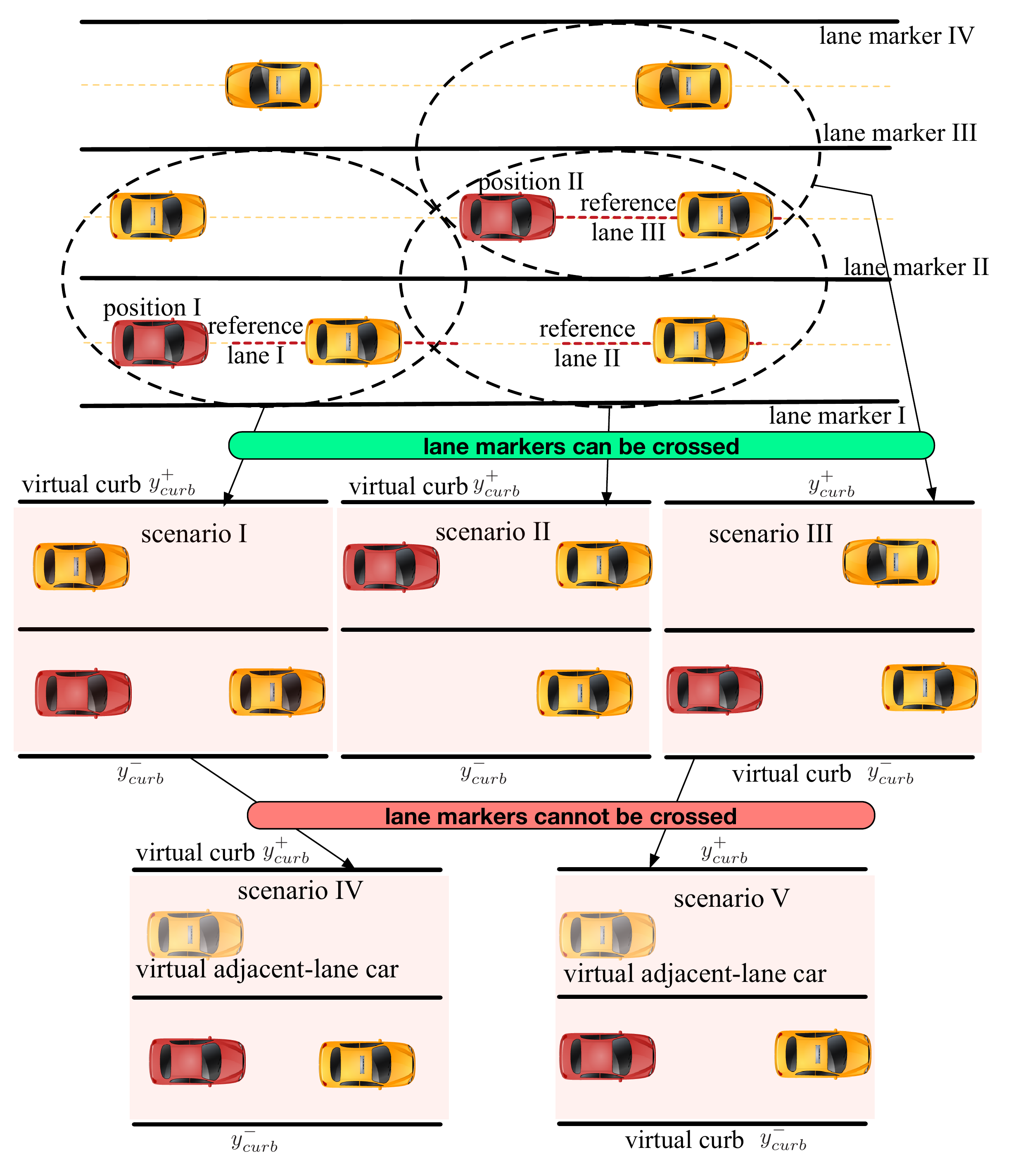}
		\par
	\end{centering}
	\caption{Generalization to multiple-lane driving scenarios\label{fig:multiple_lane}}
\end{figure}

Figure \ref{fig:multiple_lane} illustrates a scenario of multiple-lane driving. Depending on the environment settings and the driving decisions from the decision-making module, continuous driving in this scenario can be reduced to a sequence of the abstract scenarios (see Fig.~\ref{fig:long_MPC_scenario}) where the learned policy can be directly applied. 
\begin{itemize}
	\item Case 1 - both lane marker II and III can be crossed.\\
	In this case, virtual curb will be introduced to extract the abstract two-lane scenario. For instance, as shown in Fig.~\ref{fig:multiple_lane}, when the ego-vehicle is at position I, and the driving decision instructs it to remain driving on ``reference lane I'', then ``scenario I'' is extracted by setting the curbs as ``virtual curb $y^+_{curb}$'' and ``$y^-_{curb}$''. Similarly, when the ego-vehicle is at position II and the reference lane is set to be ``reference lane II'', then ``scenario II'' is used for the \textbf{\textit{policy layer}}. While if the reference lane is set to be ``reference lane III'', then ``scenario III'' is formulated by introducing the virtual lower curb $y^-_{curb}$.
	\item Case 2 - lane markers cannot be crossed.\\
	In this case, not only virtual curbs but also virtual surrounding cars are introduced to prevent the ego-vehicle from aggressive overtaking maneuvers. For instance, as shown in Fig.~\ref{fig:multiple_lane}, when the ego-vehicle is at position I and ``reference lane I'' is selected, then a virtual adjacent car moving at the same speed with the ego-vehicle is introduced for feature extraction. This transforms ``scenario I'' to ``scenario IV''. Similarly, when the ego-vehicle is at position II and ``reference lane III'' is set, ``scenario III'' changes to ``scenario V''. Introduction of virtual adjacent cars changes the features in the \textbf{\textit{policy layer}} and makes the learned policy ``think'' that it is unsafe to overtake and remain on the reference lane.
\end{itemize}

Therefore, through the combination of the decision-making module and virtual features (virtual curbs and virtual adjacent-lane cars), the learned policy can be well generalized to more practial and complicated driving scenarios.
\section{Illustrative Examples}
\label{sec:results}
In this section, several illustrative examples are given to show the effectiveness and efficiency of the proposed fast integrated planning and control framework. Performance improvement of the \textbf{\textit{policy layer}} via the Sampled-DAgger procedure is also demonstrated. For more results beyond the examples, one may refer to the complementary video at: \textit{http://iros2017autodrivinglearning.weebly.com}.

The simulation environment is established based on a 1/10 scale RC car with a sampling period of $dt{=}0.1$s. Detailed parameters can be found in Table I. All simulations are performed using Julia on a Macbook Pro (2.5 GHz Intel Core i7). Using standard Ipopt solver, the worst-case runtime for the \textbf{\textit{policy layer}} and the \textbf{\textit{execuation layer}} are, respectively, $6.42{\times10}^{{-}6}$s and $0.0766$s (Note that with warm starts, more efficient language and solvers such as C++ and SLSQP, this runtime can be further reduced). Therefore, real-time responses to rapidly changing environments are guaranteed. 
\begin{table}[h]
	\label{tab:parameter_simulation}	
	\caption{Parameters in the simulation environment}
	\centering
	\begin{tabular}{|c|c|c|c|c|}
		\hline
		$W_L$ & $L_r$ & $L_f$  & $W$ & $\epsilon$\\
		\hline
		0.38 (m)&0.19 (m) &0.21 (m) & 0.19 (m) & 0.02 (m)\\
		\hline
		 $N$ & $N_{short}$ & $V_{max}$ & $a^+_{max}$& $a^-_{max}$\\
		\hline
		 30 & 5 &1.0 (m/s) & 0.5 (m/$s^2$)& -1.0 (m/$s^2$) \\
		\hline
	\end{tabular}
\end{table}

\subsection{Overtaking Maneuver}
\begin{figure}[!h]
	\begin{centering}
		\includegraphics[clip,scale=0.4]{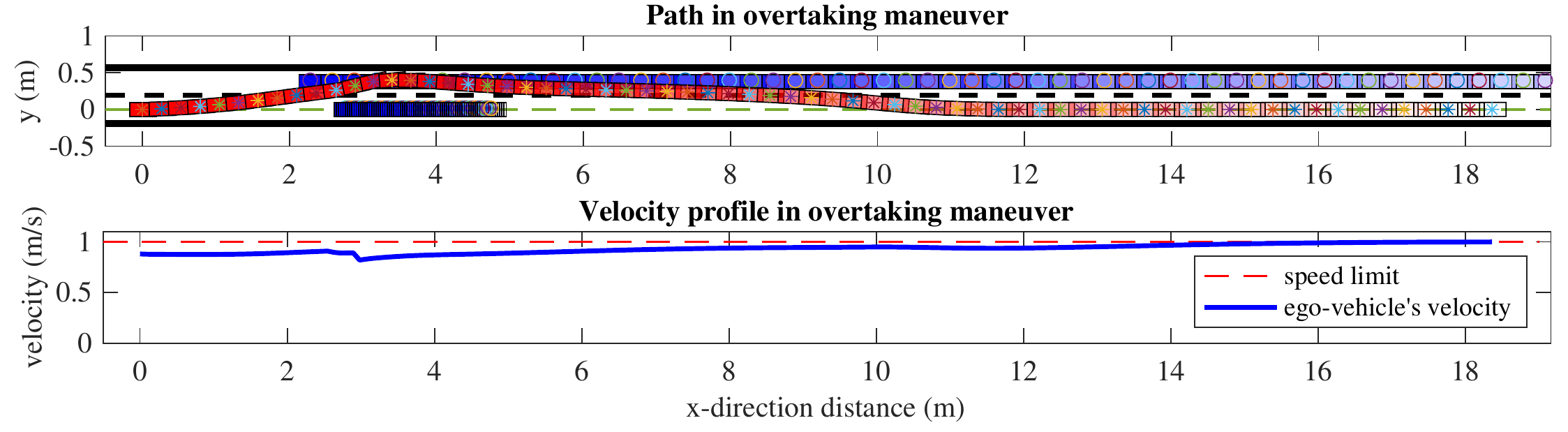}
		\par
	\end{centering}
	\caption{Overtaking maneuver using the proposed planning and control framework \label{fig:overtaking}}
\end{figure}

Figure \ref{fig:overtaking} shows the results of an overtaking maneuver, where the ego-vehicle (red) bypasses the slowly moving front car (blue) when the left lane is clear (left-lane car is also represented by blue rectangles). It can be seen that the learned policy considers long-term planning, which enables the ego-vehicle to start to accelerate and turn left at an early stage for a smoother and safe overtaking trajectory. When the yaw angle of the ego vehicle is relatively large, the learned policy brakes a bit to avoid possible collision with curbs. After that, when the environment is safe, the learned policy instructed the ego-vehicle to accelerate to speed limit and smoothly bypass the front car and return back to its original reference lane. 

\begin{figure}[!h]
	\begin{centering}
		\includegraphics[clip,scale=0.4]{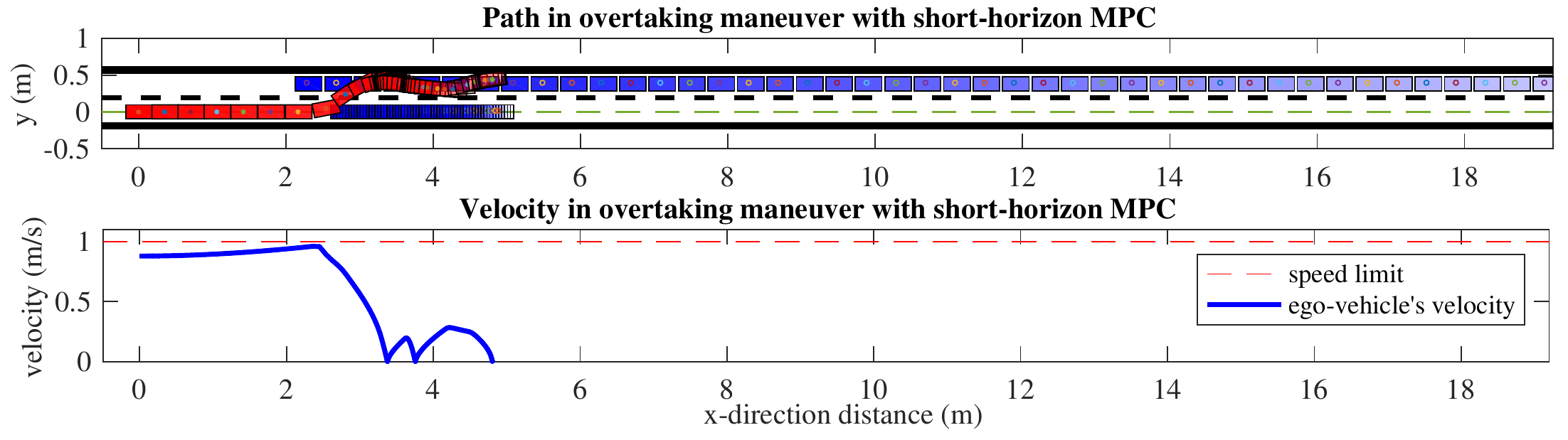}
		\par
	\end{centering}
	\caption{Overtaking maneuver using only a short-horizon MPC without the learned policy\label{fig:bad_short_MPC}}
\end{figure}

Results under the same scenario using only a short-horizon\footnote[2]{The horizon length is also set as five for comparable computation load with our proposed framework.} MPC are also shown in Fig.~\ref{fig:bad_short_MPC} as comparison. Compared to our proposed framework (see Fig.~\ref{fig:overtaking}), the ego-vehicle without the learned policy but only a short-horizon MPC drives almost blindly: acceleration without long-prepared turning, sharp brakes and large steering angles when it is too close to the front car and finally repeatedly curb hittings and stops. 
\subsection{Car Following}
Another scenario is the car-following maneuver. In such scenarios, no overtaking is allowed due to either safety (left lane is not clear) or environment settings such as no-passing solid double yellow lines. As discussed in Section \ref{sec:generalization}, to achieve this, a virtual left-lane car needs to be introduced in order to utilize the same learned policy in the \textbf{\textit{policy layer}}.

\begin{figure}[!h]
	\begin{centering}
		\includegraphics[clip,scale=0.42]{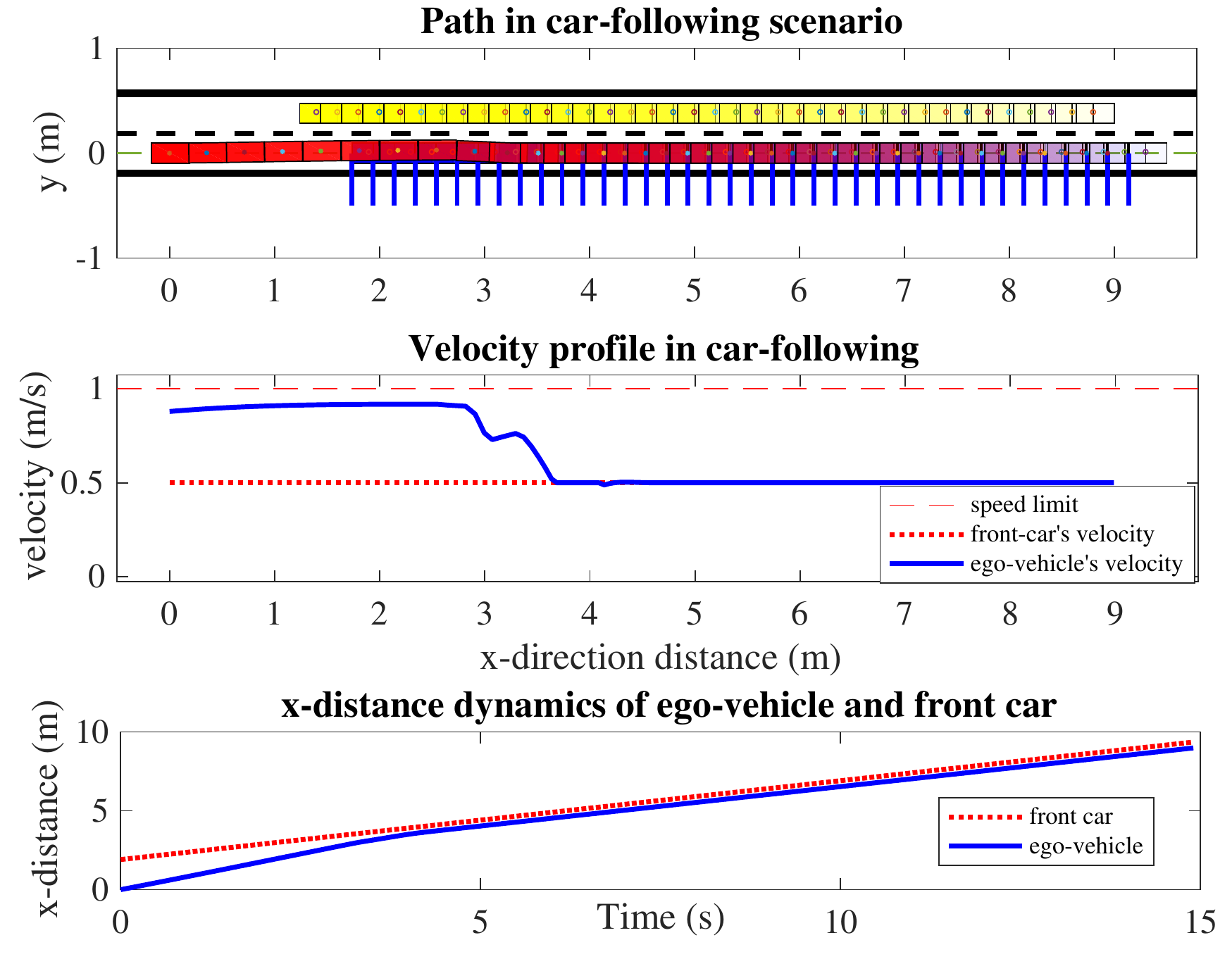}
		\par
	\end{centering}
	\caption{Car following with the proposed framework\label{fig:car_following}}
\end{figure}

The results are shown in Fig.~\ref{fig:car_following}, where the red, blue and yellow rectangles represent the ego-vehicle, the front car and the virtual car on adjacent lane, respectively. The blue bars are the rear positions of the front car. The corresponding velocity profile shows that with a fast initial speed, the ego-vehicle is forced to decelerate and perform car-following maneuver with the end-speed equal to that of the front car. Similarly, the x-direction behaviors of the ego-vehicle and the front car further verifies the collision-free car-following maneuver.

\textit{Remark III}: Notice that in Fig.~\ref{fig:car_following}, the velocity profile of the ego-vehicle is not smooth enough due to data insufficiency. Such non-smoothness can be further improved by more DAgger iterations. In Section \ref{sec:results}-C, an example will be given to show the effectiveness of the DAgger process.
\subsection{Straight Going}
\begin{figure}[!h]
	\begin{centering}
		\includegraphics[clip,scale=0.36]{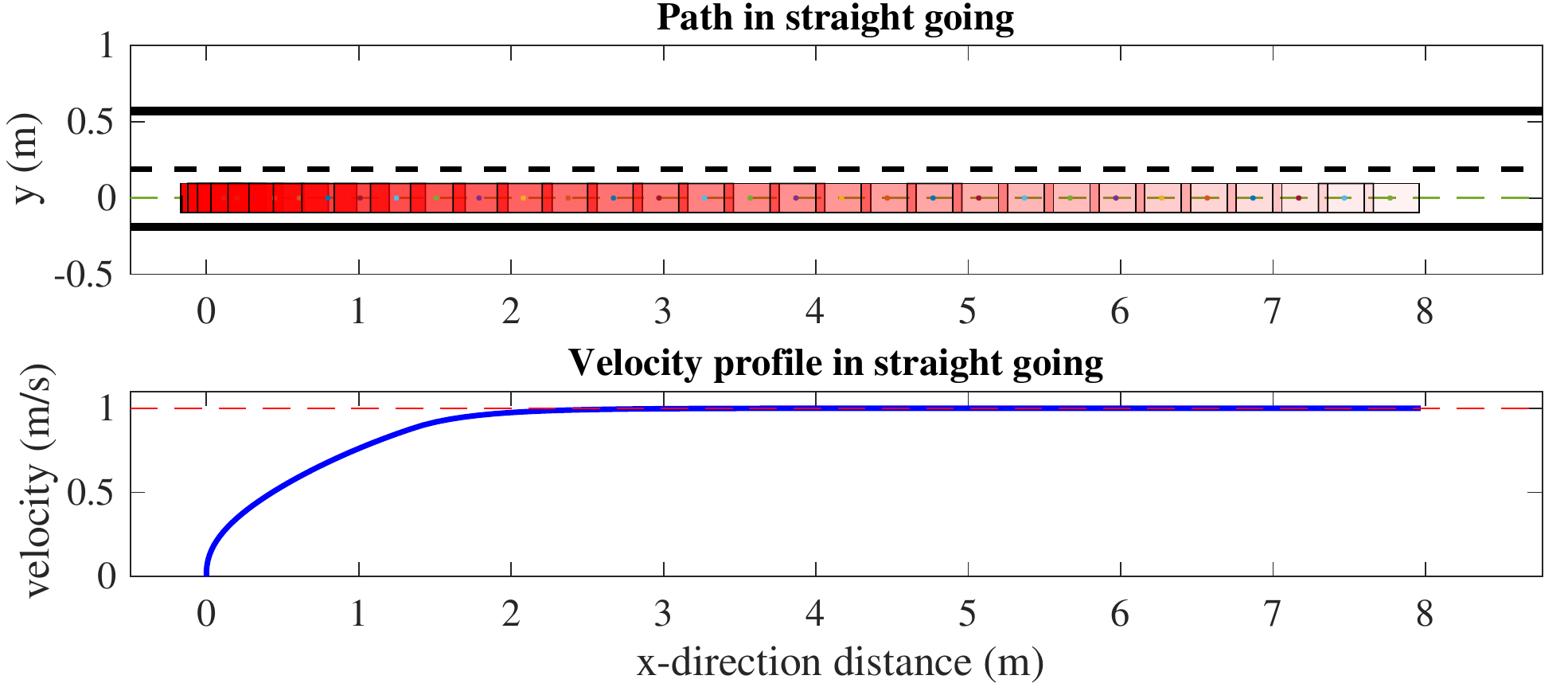}
		\par
	\end{centering}
	\caption{Straight going with the proposed framework\label{fig:straight_from_zerospeed}}
\end{figure}

Figure \ref{fig:straight_from_zerospeed} shows the results of straight-going with the proposed framework. As shown, when there is no surrounding vehicles, the learned policy is able to drive the ego-vehicle to accelerate from zero speed to speed limit without deviations from the reference lane.

\begin{figure}[!h]
	\begin{centering}
		\includegraphics[clip,width=7.0cm,scale=0.4]{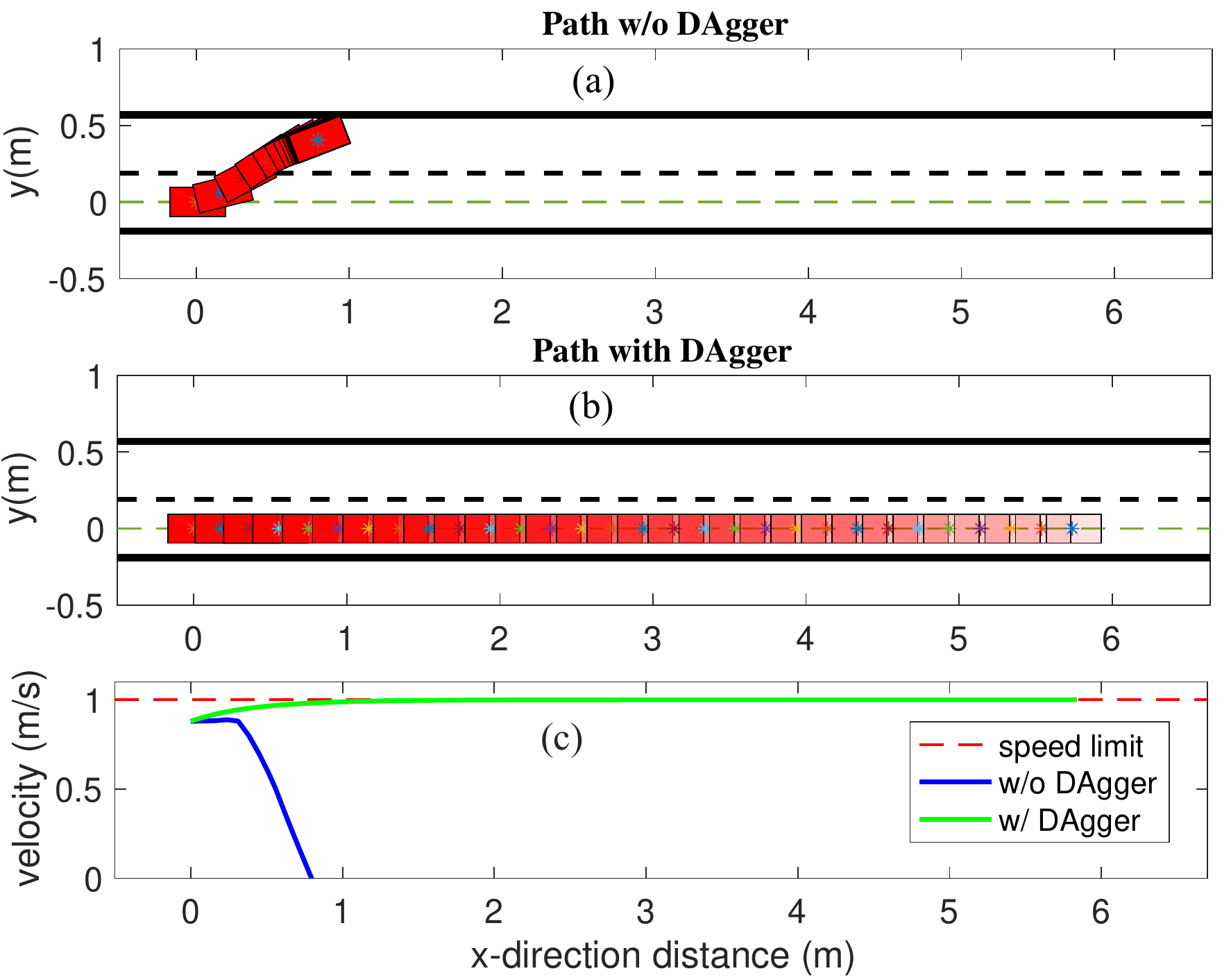}
		\par
	\end{centering}
	\caption{Learned policy improvement with customized DAgger\label{fig:straight_with_without_DAgger}}
\end{figure}

Moreover, to show the effect of the customized DAgger procedure, an example of the improved learned policy is given in Fig.~\ref{fig:straight_with_without_DAgger}, where Fig.~\ref{fig:straight_with_without_DAgger}(a) represents the on-policy running results with the initial policy $\pi_0$ on training set $D_0$. Since there are not enough straight-going scenarios in $D_0$, when $\pi_0$ is run for test, the ego-vehicle is confused by unfamiliar features and fails to go straight as we expect. The error fastly propagates and leads to collision with curbs. On the contrary, Fig.~\ref{fig:straight_with_without_DAgger} (b) shows the results with the customized DAgger procedure. It can be seen that with more on-policy running data aggregated into the training set $D$, the learned policy can effectively improve its performance. Specifically in this case, the ego-vehicle learns going straight as fast as possible when it is safe.

\section{CONCLUSIONS}
\label{sec:conclusion}
In this paper, a fast integrated planning and control framework for autonomous driving was proposed based on imitation learning and optimization. By the selection of a set of highly-representive features and customized DAgger procedure, the proposed framework can imitate the optimal driving policy given by long-term MPC, and plan safe motions extremely fast with long-term smoothness. Several example scenarios including overtaking, car-following and straight-going were given for verification.

Moreover, the learned policy can be well generalized to more complicated driving scenarios by introducing virtual features. In this work, we have discussed the usage of virtual curbs and virtual adjacent-lane cars for multiple-lane driving and car-following. For future work, more scenarios such as intersections and curvy roads will be extended. Also, the learned policy can be applied on multiple autonomous cars (a multi-agent system), and interactions among the agents can be studied.

\addtolength{\textheight}{-12cm}   





\bibliographystyle{IEEEtran}
\bibliography{IROS2017_REF, IEEEabrv}

\end{document}